\documentclass{article}

\usepackage{microtype}
\usepackage{graphicx}
\usepackage{subfigure}
\usepackage{booktabs} 

\usepackage{hyperref}
\usepackage{bm}

\usepackage[accepted]{icml2024}

\usepackage{amsmath}
\usepackage{amssymb}
\usepackage{mathtools}
\usepackage{amsthm}

\usepackage[capitalize,noabbrev]{cleveref}

\theoremstyle{plain}

\theoremstyle{definition}

\theoremstyle{remark}

\usepackage[textsize=tiny]{todonotes}
\usepackage{xspace}
\usepackage{amssymb} 
\usepackage{graphicx}
\usepackage{cleveref}
\usepackage{enumitem}
\usepackage{multicol}
\usepackage{caption}
\usepackage{multirow}
\usepackage{colortbl}
\usepackage[normalem]{ulem}
\useunder{\uline}{\ul}{}

\crefname{equation}{Eq.}{Eq.}
\crefname{algorithm}{Alg.}{Alg.}
\crefname{figure}{Fig.}{Fig.}
\crefname{table}{Tab.}{Tab.}
\crefname{section}{Sec.}{Sec.}
\crefname{chapter}{Ch.}{Ch.}

\newcommand{\modelname}{X-Oscar\xspace}
\newcommand{\moduleOneBig}{Adaptive Variational Parameter\xspace}
\newcommand{\moduleOneBigShort}{AVP\xspace}
\newcommand{\moduleTwoBig}{Avatar-aware Score Distillation Sampling\xspace}
\newcommand{\moduleTwoBigShort}{ASDS\xspace}
\newcommand{\dmtet}{\textsc{DMTet}\xspace}

\icmltitlerunning{ \modelname}

\begin{document}

\twocolumn[
\icmltitle{
\modelname: A Progressive Framework for High-quality \\ Text-guided 3D Animatable Avatar Generation
}




\icmlsetsymbol{equal}{*}

\begin{icmlauthorlist}
\icmlauthor{Yiwei Ma*}{mac}
\icmlauthor{Zhekai Lin*}{mac}
\icmlauthor{Jiayi Ji}{mac}
\icmlauthor{Yijun Fan}{mac}
\icmlauthor{Xiaoshuai Sun}{mac}
\icmlauthor{Rongrong Ji}{mac} 
\end{icmlauthorlist}

\icmlaffiliation{mac}{Key Laboratory of Multimedia Trusted Perception and Efficient Computing, Ministry of Education of China,  School of Informatics, Xiamen University, 361005, P.R. China}

\icmlcorrespondingauthor{Xiaoshuai Sun}{xssun@xmu.edu.cn}


\vskip 0.3in

{\renewcommand\twocolumn[1][]{#1}%
\begin{center}
    \vspace{-1em}
    \centering
    \captionsetup{type=figure}
    \includegraphics[width=1.0\textwidth]{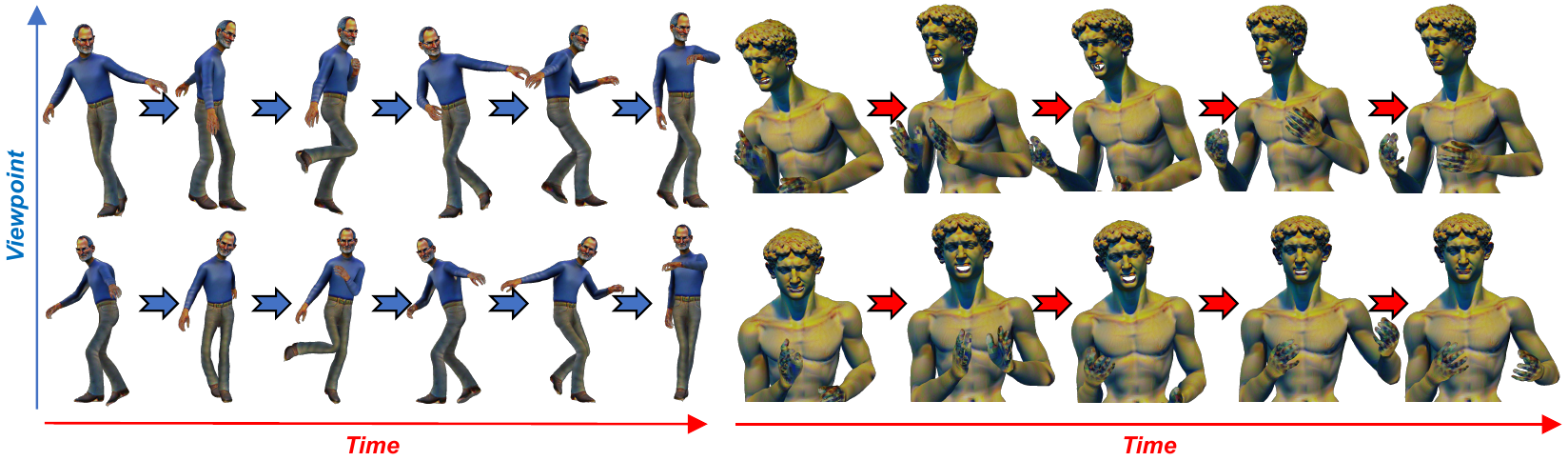}
    \vspace{-2.0em}
    \captionof{figure}{Samples generated by \modelname along temporal and viewpoint dimensions. Left Prompt: ``Steven Paul Jobs''. Right Prompt: ``David of Michelangelo''. }
    \label{fig:intro}
\end{center}}

]




\printAffiliationsAndNotice{\icmlEqualContribution} 

\begin{abstract}

Recent advancements in automatic 3D avatar generation guided by text have made significant progress.
However, existing methods have limitations such as oversaturation and low-quality output.
To address these challenges, we propose \modelname, a progressive framework for generating high-quality animatable avatars from text prompts.
It follows a sequential ``Geometry→Texture→Animation" paradigm, simplifying optimization through step-by-step generation.
To tackle oversaturation, we introduce \moduleOneBig (\moduleOneBigShort), representing avatars as an adaptive distribution during training. 
Additionally, we present \moduleTwoBig (\moduleTwoBigShort), a novel technique that incorporates avatar-aware noise into rendered images for improved generation quality during optimization.
Extensive evaluations confirm the superiority of \modelname over existing text-to-3D and text-to-avatar approaches.
Our anonymous project page: \href{https://xmu-xiaoma666.github.io/Projects/X-Oscar/}{\textcolor{red}{\underline{\textbf{https://xmu-xiaoma666.github.io/Projects/X-Oscar/}}}}.
\end{abstract}

\section{Introduction}
\label{sec:intro}

The creation of high-quality avatars holds paramount importance in a wide range of applications, including cartoon production~\cite{li2022graph,zhang2022sprite}, virtual try-on~\cite{santesteban2021self,santesteban2022ulnef}, immersive telepresence~\cite{li2020volumetric,li2020monocular,xiu2023econ}, and video game design~\cite{zheng2021deepmulticap,zhu2020reconstructing}.
Conventional methods for avatar creation are notorious for being time-consuming and labor-intensive, often demanding thousands of hours of manual work, specialized design tools, and expertise in aesthetics and 3D modeling.
In this research, we propose an innovative solution that revolutionizes the generation of high-quality 3D avatars with intricate geometry, refined appearance, and realistic animation, solely based on a text prompt. Our approach eliminates the need for manual sculpting, professional software, or extensive artistic skills, thus democratizing avatar creation and making it accessible to a broader audience.

The emergence of deep learning has brought forth a new era in 3D human body reconstruction, showcasing promising methods for automatic reconstruction from photos~\cite{liao2023high,han2023high,men2024en3d,zhang2023sifu} and videos~\cite{weng2022humannerf,jiang2022neuman}.
However, these approaches primarily focus on reconstructing human bodies from visual cues, limiting their applicability to real-world scenarios and posing challenges when it comes to incorporating creativity, editing, and control.
%
Recent advancements in large-scale vision-language models (VLM)~\cite{radford2021learning,li2022blip,li2023blip2,xu2023mplug2,ma2023towards} and diffusion models~\cite{ho2020denoising,sohl2015deep,welling2011bayesian,kulikov2023sinddm} have opened up exciting possibilities for generating 3D objects and avatars from text prompts. These methods effectively combine pretrained VLMs and diffusion models with 3D representations such as DeepSDF~\cite{park2019deepsdf}, NeRF~\cite{mildenhall2021nerf}, \dmtet~\cite{shen2021deep}, and 3D Gaussian Splatting~\cite{kerbl20233d}.
Despite these promising developments, current approaches still face several limitations. 
Some methods~\cite{ma2023xmesh,chen2023fantasia3d,wang2023prolificdreamer} focus solely on generating static everyday objects, lacking animation ability.
Other methods that aim to generate avatars based on human prior knowledge often suffer from poor geometry and appearance quality~\cite{liao2023tada,hong2022avatarclip,zhang2023avatarstudio} or are incompatible with conventional computer graphics workflows~\cite{liu2023humangaussian,huang2023dreamwaltz,cao2023dreamavatar}.

This paper presents \modelname, an innovative and advanced framework that leverages text prompts to generate high-quality animatable 3D avatars.
Specifically, \modelname builds upon the SMPL-X body model~\cite{pavlakos2019expressive} as prior knowledge and employs a strategic optimization sequence of ``Geometry → Texture → Animation".
To overcome the common challenge of oversaturation during avatar generation, we propose \moduleOneBig (\moduleOneBigShort), a novel technique that utilizes a trainable adaptive distribution to represent the geometry and appearance of the avatars.
By optimizing the distribution as a whole instead of focusing on specific parameters, \modelname effectively mitigates oversaturation, resulting in visually appealing avatars.
Furthermore, we introduce \moduleTwoBig (\moduleTwoBigShort), an innovative module that incorporates geometry-aware and appearance-aware noise into the rendered image during the optimization process.
This strategic approach significantly enhances the visual attributes of the avatars and improves their geometry and appearance quality.
Extensive experimentation demonstrates the superiority of \modelname over existing methods, showcasing improvements in both geometry and appearance quality.
Moreover, the avatars generated by \modelname are fully animatable, unlocking exciting possibilities for applications in gaming, animation, and virtual reality.

To summarize, our main contributions are three-fold:
\begin{itemize}[itemsep=1pt,topsep=0pt,parsep=0pt]
    \item We present \modelname, an innovative and progressive framework that enables the creation of delicate animatable 3D avatars from text prompts.
    \item To overcome the persistent challenge of oversaturation, we propose \moduleOneBig (\moduleOneBigShort), which represents avatars as adaptive distributions instead of specific parameters.
    \item We introduce \moduleTwoBig (\moduleTwoBigShort), an advanced module that incorporates geometry-aware and appearance-aware noise into the rendered image during the optimization process, resulting in high-quality outputs.
\end{itemize}

\section{Related Work}
\label{sec:relate}

\textbf{Text-to-3D Generation.}
The emergence of vision-language models (VLMs)~\cite{radford2021learning,ma2022xclip} and diffusion models has brought about a revolutionary impact on text-to-3D content generation.
Pioneering studies like CLIP-forge~\cite{sanghi2022clip}, DreamFields~\cite{jain2022zero}, CLIP-Mesh~\cite{mohammad2022clip}, and XMesh~\cite{ma2023xmesh} have showcased the potential of utilizing CLIP scores~\cite{radford2021learning} to align 3D representations with textual prompts, enabling the generation of 3D assets based on textual descriptions.
Subsequently, DreamFusion~\cite{poole2022dreamfusion} introduced Score Distillation Sampling (SDS), a groundbreaking technique that leverages pretrained diffusion models~\cite{saharia2022photorealistic} to supervise text-to-3D generation. This approach has significantly elevated the quality of generated 3D content.
Building on these foundations, researchers have explored various strategies to further enhance text-to-3D generation. These strategies encompass coarse-to-fine optimization~\cite{lin2023magic3d}, conditional control~\cite{li2023mvcontrol,chen2023control3d}, bridging the gap between 2D and 3D~\cite{ma2023xdreamer}, introducing variational score distillation~\cite{wang2023prolificdreamer}, and utilizing 3D Gaussian Splatting~\cite{chen2023text,li2023gaussiandiffusion,yi2023gaussiandreamer,tang2023dreamgaussian}.
Nevertheless, despite these advancements, existing methodologies primarily concentrate on generating common static objects. When applied to avatar generation, they face challenges such as poor quality and the inability to animate the generated avatars.
In contrast, our proposed framework, \modelname, specifically aims to generate high-quality 3D animatable avatars from text prompts. \modelname caters to the unique requirements of avatar generation, including intricate geometry, realistic textures, and fluid animations, to produce visually appealing avatars suitable for animation.

\textbf{Text-to-Avatar Generation.}
The domain of text-to-avatar generation~\cite{kolotouros2024dreamhuman,zhang2024avatarverse,huang2023humannorm,xu2023seeavatar,zhou2024headstudio} has emerged as a prominent and vital research area to cater to the demands of animated avatar creation.
This field incorporates human priors such as SMPL~\cite{loper2023smpl}, SMPL-X~\cite{pavlakos2019SMPLX}, and imGHUM~\cite{alldieck2021imghum} models.
AvatarCLIP~\cite{hong2022avatarclip} utilizes SMPL and Neus~\cite{wang2021neus} models to generate 3D avatars guided by the supervision of CLIP scores.
Dreamwaltz~\cite{huang2023dreamwaltz} introduces NeRF~\cite{mildenhall2021nerf} to generate 3D avatars based on 3D-consistent occlusion-aware SDS and 3D-aware skeleton conditioning.
AvatarBooth~\cite{zeng2023avatarbooth} leverages dual fine-tuned diffusion models to achieve customizable 3D human avatar generation.
AvatarVerse~\cite{zhang2023avatarverse} utilizes ControlNet~\cite{zhang2023adding} and DensePose~\cite{guler2018densepose} to enhance view consistency.
TADA~\cite{liao2023tada} employs a displacement layer and a texture map to predict the geometry and appearance of avatars.
HumanNorm~\cite{huang2023humannorm} proposes a normal diffusion model for improved geometry.
HumanGaussian~\cite{liu2023humangaussian} uses 3D Gaussian Splatting as human representation for text-to-avatar generation.
Despite these advancements, existing methods often produce low-quality and over-saturated results. 
To overcome these limitations, we introduce a progressive framework that incorporates two key modules, namely \moduleOneBig and \moduleTwoBig. Our framework effectively generates high-fidelity avatars that are visually appealing and realistic.

\section{Preliminaries}

\textbf{Score Distillation Sampling (SDS)}~\cite{poole2022dreamfusion}, also known as Score Jacobian Chaining (SJC)~\cite{wang2023score}, is a powerful optimization method that adapts pretrained text-to-image diffusion models for text-to-3D generation.
Given a pretrained diffusion model $p_\phi({z}_t | y, t)$, where $\phi$ represents the model's parameters, $y$ is the input text prompt, and ${z}_t$ denotes the noised image at timestep $t$, SDS aims to optimize a 3D representation to align with the text prompt.
The forward diffusion process in SDS is formulated as $q({z}_t |g(\theta,c), y, t)$, where $\theta$ represents the trainable parameters of the 3D representation, $c$ denotes the camera, and $g(\cdot)$ is the rendering function.
The objective of SDS can be expressed as follows:
\begin{equation}
\begin{aligned}
    &\operatorname*{min}{\cal L}_{\mathrm{SDS}}(\theta)= \\
    &\mathbb{E}_{(t,c)}\left[\sqrt{\frac{1-\gamma_{t}}{\gamma_{t}}}\omega(t){\cal D}_{\mathrm{KL}}(
    q({z}_t |g(\theta,c), y, t)
    \parallel 
    p_\phi({z}_t | y, t)
    )\right],
\end{aligned}
\label{eq:sds}
\end{equation}
where $\omega(t)$ is a weighting function dependent on the timestep $t$, $z_t = \sqrt{\gamma_t} g(\theta,c) + \sqrt{1-\gamma_t}\epsilon$ is the noised image, and ${\cal D}_\mathrm{KL}(\cdot)$ represents the Kullback-Leibler Divergence~\cite{kullback1951information}. 
To approximate the gradient of the SDS objective, the following equation is leveraged:
\begin{equation}
\begin{small}
    \begin{aligned}
    \nabla_{\theta}{\mathcal{L}}_{\mathrm{{SDS}}}(\theta)\triangleq\mathbb{E}_{t,\epsilon,c}\,\left[\omega(t)(\underbrace{\hat{{\epsilon}}_\phi({z}_{t};y,t)}_{\text{predicted noise}}-\underbrace{\epsilon}_{\text{Guassian noise}}){\frac{\partial g(\theta,c)}{\partial\theta}}\right],
\end{aligned}
\end{small}
\label{eq:g_sds}
\end{equation}
where ${\epsilon}\sim {\mathcal{N}}\left({0},{{ I}}\right) $ represents sampled noise from a normal distribution, and $\hat{{\epsilon}}_\phi({z}_{t};y,t)$ denotes the predicted noise of the pretrained diffusion model at timestep $t$.

\textbf{SMPL-X}~\cite{pavlakos2019SMPLX} is a widely adopted parametric 3D human body model in the fields of computer graphics and animation.
It offers a comprehensive representation of the human body, consisting of $10,475$ vertices and $54$ joints, facilitating detailed and realistic character rendering.
By specifying shape $\mathfrak{s}$, pose $\mathfrak{p}$, and expression $\mathfrak{e}$ parameters, the SMPL-X model generates a human body using the following equation:
\begin{equation}
\begin{aligned}
{{\mathrm{T}(\mathfrak{s},\mathfrak{p},\mathfrak{e})=\mathcal{T}+B_{s}(\mathfrak{s})+B_{p}(\mathfrak{p})+B_{e}(\mathfrak{e}),}}
\end{aligned}
\label{eq:smplx1}
\end{equation}
where $\mathcal{T}$ denotes a standard human template, while $B_{s}(\cdot), B_{p}(\cdot), B_{e}(\cdot)$ represent shape, expression, and pose blend shapes, respectively.
These blend shapes deform the template to generate a wide range of body shapes, poses, and expressions.
To transition the human body from a standard pose to a target pose, linear blend skinning (LBS) is employed:
\begin{equation}
{\mathrm{M}(\mathfrak{s},\mathfrak{p},\mathfrak{e})={\mathcal{W}_{LBS}}(\mathrm{T}(\mathfrak{s},\mathfrak{p},\mathfrak{e}),J(\mathfrak{s}),\mathfrak{p},{W})},
\label{eq:smplx2}
\end{equation}
where $\mathcal{W}_{LBS}(\cdot)$ represents the LBS function, $J(\mathfrak{s})$ corresponds to the skeleton joints, and ${W}$ represents the skinning weight.
The LBS function calculates the final vertex positions by interpolating between the deformed template vertices based on the assigned skinning weights. 
This process ensures a smooth and natural deformation of the body mesh.

\begin{figure*}
\centering
\includegraphics[width=2.1\columnwidth]{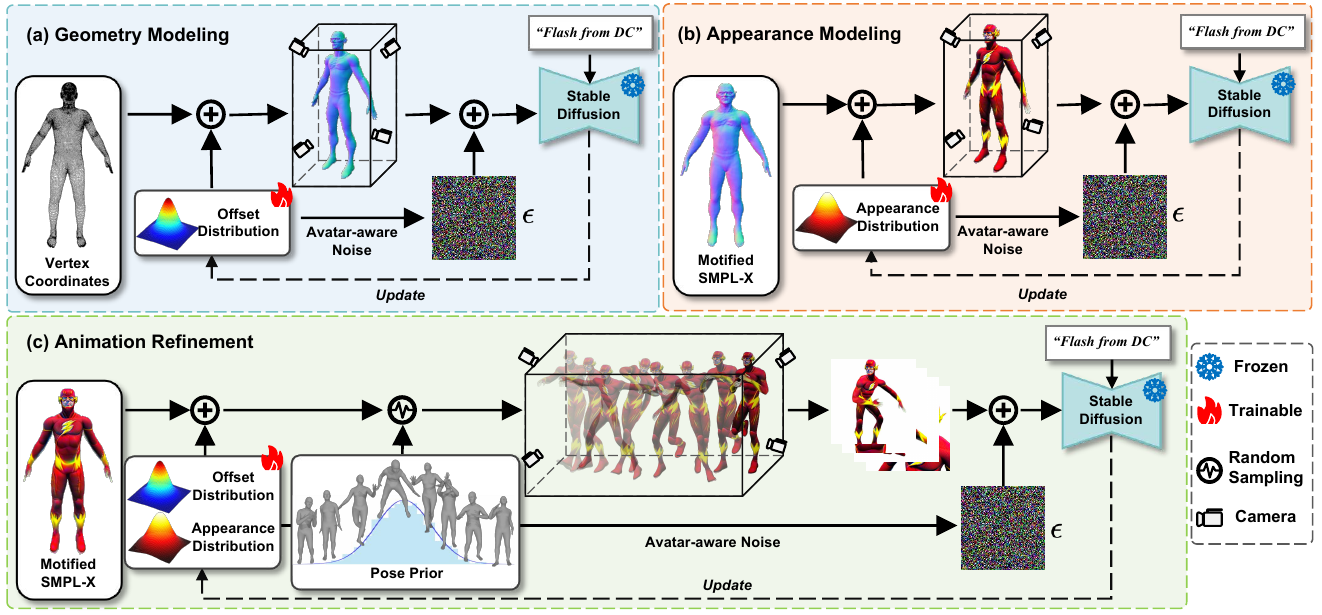}
\vspace{-1.5em}
\caption{Overview of the proposed \modelname, which consists of three generation stages: (a) geometry modeling, (b) appearance modeling, and (c) animation refinement. }
\label{fig:oveview}
\vspace{-1.5em}
\end{figure*}

\section{Approach}
\label{sec:approach}

\begin{figure}[]
\centering
\includegraphics[width=0.9\columnwidth]{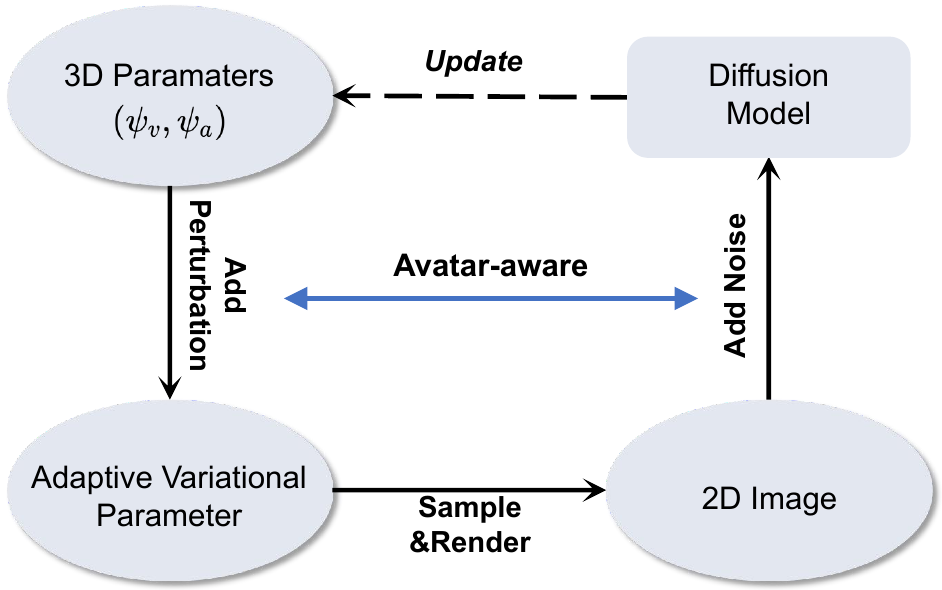}
\vspace{-0.5em}
\caption{The workflow of the proposed \modelname. 
First, we incorporate the adaptive perturbation into the 3D parameters, forming the avatar distribution.
Next, we sample a set of parameters from the avatar distribution and render a 2D image.
Finally,  we apply avatar-aware noise to the rendered image for denoising to optimize 3D parameters.
}
\label{fig:workflow}
\vspace{-1.0em}
\end{figure}

The overview of \modelname is depicted in \cref{fig:oveview}, and the workflow is illustrated in \cref{fig:workflow}.
In the upcoming sections, we present a comprehensive description of the \modelname framework:
In \cref{sec:progressive}, we delve into the progressive modeling pipeline of \modelname. This pipeline breaks down the complex task of avatar generation into three manageable subtasks, with each subtask focusing on a specific aspect of avatar creation.
In \cref{sec:moduleone}, we introduce \moduleOneBig (\moduleOneBigShort). This component employs a trainable adaptive distribution to represent the avatar, addressing the issue of oversaturation that is commonly encountered in avatar generation.
In \cref{sec:moduletwo}, we present \moduleTwoBig (\moduleTwoBigShort). This module incorporates geometry-aware and appearance-aware noise into the denoising process, enabling the pretrained diffusion model to perceive the current state of the generated avatar, resulting in the production of high-quality outputs.
%


\subsection{Progressive Modeling}
\label{sec:progressive}

\textbf{Geomotry Modeling.}
During this phase, our objective is to optimize the geometry of the avatars, represented by the SMPL-X model, to align with the input text prompt $y$.
Formally, we aim to optimize the trainable vertex offsets $\psi_v \in \mathbb{R}^{N \times 3}$, initialized as a matrix of zeros, to align the modified vertex coordinates $\nu^\prime = \nu + \psi_v$ with the text prompt $y$, where $\nu$ represents the vertex coordinates of the template avatar body, and $N$ is the number of vertices of the SMPL-X model.
To achieve this, we utilize a differentiable rendering pipeline. By taking the original mesh $\mathcal{M}$ of SMPL-X and the predicted vertex offsets $\psi_v$ as inputs, we render a normal image $\mathcal{N}$ of the modified mesh using a differentiable renderer~\cite{laine2020modular}:
\begin{equation}
\mathcal{N} = g(\mathcal{M},\psi_v,c),
\label{eq:normal}
\end{equation}
where $g(\cdot)$ denotes the rendering function, and $c$ represents a randomly sampled camera parameter.
In each iteration, we introduce Gaussian noise $\epsilon$ to the normal map $\mathcal{N}$ and apply a pretrained Stable Diffusion (SD) model~\cite{rombach2022high} to denoise it. The gradient of the trainable vertex offsets $\psi_v$ during denoising is then calculated as follows:
\begin{footnotesize}
\begin{equation}    
    \nabla_{\psi_v}\mathcal{L}_{\mathrm{geo}}(\psi_v,\mathcal{N})=\mathbb{E}_{t,\epsilon}\left[w(t)\left(\hat{\epsilon}_{\phi}(z_{t}^{\mathcal{N}};y,t)-\epsilon\right)\frac{\partial\mathcal{N}}{\partial\psi_v}\right],
\label{eq:normal_sds}
\end{equation}
\end{footnotesize}
where $\hat{\epsilon}_{\phi}(z_{t}^{\mathcal{N}};y,t)$ represents the predicted noise by SD based on the timestep $t$, input text embedding $y$, and the noisy normal image $z_t^\mathcal{N}$.

\textbf{Appearance Modeling.}
After completing the geometry modeling phase, we obtain a mesh that aligns with the prompt in terms of shape, with vertex coordinates $\nu^\prime = \nu + \psi_v$.
In this stage, our objective is to optimize an albedo map $\psi_a \in \mathbb{R}^{h \times w \times 3}$ to represent the appearance of the resulting avatar, where $h$ and $w$ represent the height and width of the albedo map.
To achieve this, we start by rendering a colored image $\mathcal{I}$ from a randomly sampled camera parameter $c$ based on the vertex offsets $\psi_{v}$ and the albedo map $\psi_{a}$ using a differentiable renderer~\cite{laine2020modular}:
\begin{equation}
\mathcal{I} = g(\mathcal{M},\psi_v,\psi_a,c).
\label{eq:color}
\end{equation}
To optimize the albedo map $\psi_a$, we employ a loss function similar to \cref{eq:normal_sds} used in the geometry modeling phase:
\begin{footnotesize}
\begin{equation}    
    \nabla_{\psi_a}\mathcal{L}_{\mathrm{app}}(\psi_a,\mathcal{I})=\mathbb{E}_{t,\epsilon}\left[w(t)\left(\hat{\epsilon}_{\phi}(z_{t}^{\mathcal{I}};y,t)-\epsilon\right)\frac{\partial\mathcal{I}}{\partial\psi_a}\right],
\label{eq:color_sds}
\end{equation}
\end{footnotesize}
where $\hat{\epsilon}_{\phi}(z_{t}^{\mathcal{I}};y,t)$ represents the predicted noise by the SD model.
This loss function encourages the rendered image $\mathcal{I}$ to align with the text prompt $y$ by minimizing the discrepancy between the predicted noise $\hat{\epsilon}_{\phi}$ and the added Gaussian noise $\epsilon$.
By optimizing the albedo map $\psi_a$ using this loss function, we can generate appearances for the avatars that are consistent with the provided text prompts.

\textbf{Animation Refinement.}
Given that both the geometry modeling and appearance modeling stages optimize the avatar in a canonical pose, it is inevitable that certain parts of the avatar may be obstructed, leading to lower-quality results in those areas.
To overcome this challenge, we introduce an animation refinement stage where we adjust the pose of the avatar and simultaneously optimize both the geometry and appearance.
Specifically, we sample viable pose parameters $\mathfrak{p}$ from a pre-trained model such as VPoser~\cite{pavlakos2019expressive}.
For each sampled pose, we render the normal image $\mathcal{N}_p$ and colored image $\mathcal{I}_p$ of the animated avatar using a differentiable renderer~\cite{laine2020modular}:
\begin{equation}
\mathcal{N}_p = g(\mathcal{M},\psi_v,c,\mathfrak{p}), \quad
\mathcal{I}_p = g(\mathcal{M},\psi_v,\psi_a,c,\mathfrak{p}),
\label{eq:animation}
\end{equation}
where pose parameters $\mathfrak{p}$ and camera parameters $c$ vary in each iteration.
To optimize the geometry and appearance of the avatar in the animated pose, we define an animation loss $\mathcal{L}_{\mathrm{ani}}$ as follows:
\begin{equation}
\mathcal{L}_{\mathrm{ani}}(\psi_v,\psi_a,\mathcal{N}_p,\mathcal{I}_p) = \mathcal{L}_{\mathrm{geo}}(\psi_v,\mathcal{N}_p) + \mathcal{L}_{\mathrm{app}}(\psi_v,\psi_a,\mathcal{I}_p),
\label{eq:animation_loss}
\end{equation}
where $\mathcal{L}_{\mathrm{geo}}$ and $\mathcal{L}_{\mathrm{app}}$ are the geometry loss and appearance loss, respectively.
The gradients of the animation loss for the vertex offsets $\psi_v$ and the albedo maps $\psi_a$ are calculated as follows:
\begin{scriptsize}
\begin{equation}    
\begin{aligned}
     &\nabla_{\psi_v}\mathcal{L}_{\mathrm{ani}}(\psi_v,\mathcal{N}_p,\mathcal{I}_p) \\
    =&\mathbb{E}_{t,\epsilon}  \left[
    w(t)\left(\hat{\epsilon}_{\phi}(z_{t}^{\mathcal{N}_p};y,t)-\epsilon\right)\frac{\partial\mathcal{N}_p}{\partial\psi_v}
    +
    w(t)\left(\hat{\epsilon}_{\phi}(z_{t}^{\mathcal{I}_p};y,t)-\epsilon\right)\frac{\partial\mathcal{I}_p}{\partial\psi_v}
    \right],
\end{aligned}
\label{eq:animation_sds1}
\end{equation}
\end{scriptsize}
\begin{scriptsize}
\begin{equation}    
\begin{aligned}
    &\nabla_{\psi_a}\mathcal{L}_{\mathrm{ani}}(\psi_a,\mathcal{I}_p)=\mathbb{E}_{(t,\epsilon)}\left[w(t)\left(\hat{\epsilon}_{\phi}(z_{t}^{\mathcal{I}_p};y,t)-\epsilon\right)\frac{\partial\mathcal{I}_p}{\partial\psi_a}\right],
\end{aligned}
\label{eq:animation_sds2}
\end{equation}
\end{scriptsize}
The notations used here are similar to those defined in \cref{eq:g_sds}.
By minimizing the animation loss using these gradients, we refine the geometry and appearance of the avatar in various poses, resulting in improved quality in the final output.

\subsection{\moduleOneBig}
\label{sec:moduleone}
As formulated in \cref{eq:sds} and \cref{eq:g_sds}, SDS aims to optimize a precise 3D representation to align all images rendered from arbitrary viewpoints with the input prompt evaluated by 2D diffusion models.
However, there exists a fundamental contradiction between achieving an accurate 3D representation and the inherent multi-view inconsistency associated with 2D diffusion models.
Specifically, it is often unreasonable to expect high similarity scores of a 2D diffusion model between all multi-view images of a specific 3D representation and text prompts. 
%
%
Consequently, when SDS is employed to enforce similarity between each perspective of a specific 3D representation and the text prompt, it can lead to the undesirable issue of oversaturation.
To address this concern, we propose formulating the 3D representation as a distribution of vertex offsets, denoted as \textit{offset distribution}, and a distribution of albedo maps, referred to as \textit{appearance distribution}.
Specifically, we perturb $\psi_v$ and $\psi_a$ of the 3D human representation with Gaussian noises to improve the robustness of the model and alleviate the oversaturation problem. This perturbation process can be expressed as:
\begin{equation}    
\psi_v^\prime \sim \psi_v + \lambda_v{\mathcal{N}}\left({0},{{ I}}\right), \quad
\psi_a^\prime \sim \psi_a + \lambda_a{\mathcal{N}}\left({0},{{ I}}\right),
\label{eq:avp1}
\end{equation}
where $\lambda_v$ and $\lambda_a$ serve as weights to control the magnitude of the perturbations.
The mean of the offset distribution and appearance distribution can be learned by optimizing $\psi_v$ and $\psi_a$, while their standard deviations are determined by $\lambda_v$ and $\lambda_a$.
Thus, choosing appropriate values for $\lambda_v$ and $\lambda_a$ is crucial and challenging.
If these values are too small, the model may not fully benefit from learning the distributions. In extreme cases, when $\lambda_v = \lambda_a = 0$, the model essentially learns specific parameters instead of distributions.
Conversely, when $\lambda_v$ and $\lambda_a$ are excessively large, the learning process becomes challenging due to highly unstable perturbations. In extreme cases, when $\lambda_v = \lambda_a = +\infty$, the generated results become independent of the underlying $\psi_v$ and $\psi_a$.

To overcome the above challenges and facilitate a learning process that progresses from easy to difficult without manual weight assignment, we propose \moduleOneBig (\moduleOneBigShort) for 3D representation.
Specifically, we leverage the standard deviations of $\psi_v$ and $\psi_a$ as weights for perturbations, which can be formulated as follows:
\begin{equation}    
\psi_v^\prime \sim \psi_v + {\sigma(\psi_v)}{\mathcal{N}}\left({0},{{ I}}\right) = {\mathcal{N}}\left(\psi_v,\sigma(\psi_v)^2\right), 
\label{eq:avp2}
\end{equation}
\begin{equation}    
\psi_a^\prime \sim \psi_a + {\sigma(\psi_a)}{\mathcal{N}}\left({0},{{ I}}\right) = {\mathcal{N}}\left(\psi_a,\sigma(\psi_a)^2\right), 
\label{eq:avp3}
\end{equation}
where ${\sigma(\cdot)}$ represents the standard deviation.
This adaptive approach has several advantages.
\emph{Firstly, it enables the model to learn progressively from easy to difficult scenarios.}
Initially, $\psi_v$ and $\psi_a$ are initialized as matrices of all zeros and all 0.5, respectively, resulting in a standard deviation of 0. Consequently, during the early stages of training, the model focuses on optimizing the means of $\psi_v^\prime$ and $\psi_a^\prime$ to reasonable values.
As training progresses, the standard deviations gradually increase, promoting the model's ability to maintain high similarity between the 3D representation and the text even in the presence of noise interference.
\emph{Secondly, this approach is fully automatic.} 
The model learns to adapt the perturbation weights based on the current state of the 3D representation, eliminating the need for manual intervention or hyperparameter tuning.
During the inference phase, we utilize the mean values of $\psi_v^\prime$ and $\psi_a^\prime$ to represent the avatar.

\subsection{\moduleTwoBig}
\label{sec:moduletwo}

In previous work on SDS~\cite{poole2022dreamfusion}, a Gaussian noise related to timestep $t$ was introduced to the rendered image, and a pretrained diffusion model was utilized to denoise the noisy image for optimizing the 3D representation. The process of adding noise can be formulated as follows:
\begin{equation}
\begin{aligned}
    z_t 
    = &\sqrt{\alpha_t} z_{t-1} + \sqrt{1-\alpha_t} \epsilon_{t-1} \\
    = &\sqrt{\alpha_t \alpha_{t-1}} z_{t-2} + \sqrt{1-\alpha_t \alpha_{t-1}} \bar{\epsilon}_{t-2} \\
    = &\cdots \\
    = &\sqrt{\bar{\alpha}_t} z_0 + \sqrt{1-\bar{\alpha}_t} \bar{\epsilon}_0,
\label{eq:noise}
\end{aligned}
\end{equation}
where $z_t$ represents the noised image at timestep $t$, $\bar{\alpha}_{t}=\prod_{i=1}^{t}\alpha_{i}$, and $\epsilon_i, \bar{\epsilon}_i\sim{\mathcal{N}}\left({0},{{ I}}\right)$.
%
%
Since $t \sim \mathcal{U}(0.02, 0.98)$ is randomly sampled, the noise added to the rendered image is independent of the avatar's current state.
To establish a correlation between the denoising process and the avatar's current state, and to facilitate a learning process from easy to difficult, we propose \moduleTwoBig (\moduleTwoBigShort).
Specifically, the noised image with avatar-aware noise can be formulated as follows:
\begin{small}    
\begin{equation}
\begin{aligned}
    z_t 
    =&\sqrt{\bar{\alpha}} z_0 + \sqrt{1-\bar{\alpha}_t}(\lambda_n\epsilon_n+\lambda_v{\sigma(\psi_v)}\epsilon_v+\lambda_a{\sigma(\psi_a)}\epsilon_a) \\
    =&\sqrt{\bar{\alpha}} z_0 +  \sqrt{1-\bar{\alpha}_t}\sqrt{(\lambda_n)^2 + (\lambda_v\sigma(\psi_v) )^2+ (\lambda_a\sigma(\psi_a))^2} \epsilon  \\
    =& \sqrt{\bar{\alpha}} z_0 +  \sqrt{1-\bar{\alpha}_t}\epsilon_\theta ,
\label{eq:a_sds_noise}
\end{aligned}
\end{equation}
\end{small}
where $\epsilon_n$, $\epsilon_v$, $\epsilon_a$, and $\epsilon$ are i.i.d. Gaussian random variables with zero mean and unit variance, i.e., $\epsilon_n, \epsilon_v, \epsilon_a, \epsilon \sim \mathcal{N}(0, I)$, and $\epsilon_\theta \sim \mathcal{N}(0, (\lambda_n)^2 + (\lambda_v\sigma(\psi_v))^2 + (\lambda_a\sigma(\psi_a))^2)$.
At the initial stage, when $\sigma(\psi_v)=\sigma(\psi_a)=0$, the initial variance of the noise is relatively small, resulting in an easier denoising process for diffusion models.
As the training progresses, $\sigma(\psi_v)$ and $\sigma(\psi_a)$ gradually increase, leading to an increase in the noise variance.
Consequently, this increases the difficulty of denoising.
By incorporating avatar-aware noise, the model can undergo a learning process from easy to difficult.
The gradient of \moduleTwoBigShort is then formulated as follows:
\begin{equation}
\begin{aligned}
    &\nabla_{\theta}{\mathcal{L}}_{\mathrm{{ASDS}}}(\theta)\triangleq \\
    &\mathbb{E}_{(t,\epsilon,c)}\,\left[\omega(t)\big(\underbrace{\hat{{\epsilon}}_\phi({z}_{t};y,t)}_{\text{precited noise}}-\underbrace{\epsilon_\theta}_{\text{avatar-aware noise}}\big){\frac{\partial g(\theta,c)}{\partial\theta}}\right],
\end{aligned}
\label{eq:asds}
\end{equation}
where $z_t = \sqrt{\bar{\alpha}}g(\theta,c) + \sqrt{1-\bar{\alpha}}\epsilon_\theta$ represents the noised image, and $\epsilon_\theta$ is an avatar-aware noise that encourages the paradigm of learning from easy to difficult.
%

\begin{table*}[]
\centering
\caption{Quantitative comparison of SOTA Methods: The top-performing and second-best results are highlighted in \textbf{bolded} and \underline{underlined}, respectively. As AvatarCLIP employs the CLIP score as its training supervision signal, it is inappropriate to gauge its performance using the CLIP score. Therefore, we set the CLIP score of AvatarCLIP to {\color[HTML]{BFBFBF} gray}.}
\vspace{-0.9em}
\setlength{\tabcolsep}{1.2mm}{
\begin{tabular}{l|ccc|ccc|ccc}
\hline
                                  & \multicolumn{3}{c|}{\textbf{User Study}}      & \multicolumn{3}{c|}{\textbf{CLIP Score}}                                                   & \multicolumn{3}{c}{\textbf{OpenCLIP Score}}                                                \\ \cline{2-10} 
\multirow{-2}{*}{\textbf{Method}} & Geo. Qua.     & Tex. Qua.     & Tex. Con.     & ViT-B/32                     & ViT-B/16                     & ViT-L/14                     & ViT-B/32                     & ViT-B/16                     & ViT-L/14                     \\ \hline
DreamFusion                       & 2.66          & 4.18          & 3.29          & 29.29                        & 29.29                        & 25.30                        & 31.57                        & 28.22                        & 30.17                        \\
Magic3D                           & 4.21          & 3.12          & 1.61          & 28.52                        & 30.92                        & 27.02                        & 31.14                        & 28.21                        & 30.21                        \\
Fantasia3D                        & 2.14          & 2.42          & 2.53          & 30.34                        & 30.42                        & 26.12                        & 29.68                        & 28.46                        & \textbf{31.46}               \\
ProlificDreamer                   & 2.11          & 3.72          & 6.29          & 30.30                        & 30.28                        & 25.00                        & {\ul 30.81}                  & 28.59                        & 30.75                        \\
AvatarCLIP                        & 3.28          & 2.64          & 2.09          & {\color[HTML]{BFBFBF} 34.49} & {\color[HTML]{BFBFBF} 32.45} & {\color[HTML]{BFBFBF} 28.20} & {\color[HTML]{BFBFBF} 32.77} & {\color[HTML]{BFBFBF} 31.20} & {\color[HTML]{BFBFBF} 31.98} \\
AvatarCraft                       & 4.39          & 4.55          & 3.37          & 27.59                        & 29.70                        & 25.23                        & 26.19                        & 24.60                        & 25.55                        \\
DreamWaltz                        & {\ul 6.38}    & 6.09          & 6.99          & 30.86                        & {\ul 31.20}                  & 27.32                        & 30.65                        & 29.09                        & 29.83                        \\
HumanGuassian                     & 6.03          & 4.51          & 6.08          & 28.46                        & 29.18                        & 26.26                        & 26.37                        & 26.82                        & 29.09                        \\ 
TADA                              & 5.03          & {\ul 6.95}    & {\ul 7.62}    & {\ul 31.09}                  & 30.48                        & {\ul 27.72}                  & 30.67                        & {\ul 30.05}                        & 30.17                        \\ \hline
X-Oscar                           & \textbf{8.85} & \textbf{8.91} & \textbf{9.22} & \textbf{31.70}               & \textbf{31.97}               & \textbf{28.10}               & \textbf{30.91}               & \textbf{30.28}               & {\ul 30.42}                  \\ \hline
\end{tabular}
}
\label{tab:compare}
\vspace{-0.5em}
\end{table*}

\begin{figure*}[]
\centering
\includegraphics[width=1.9\columnwidth]{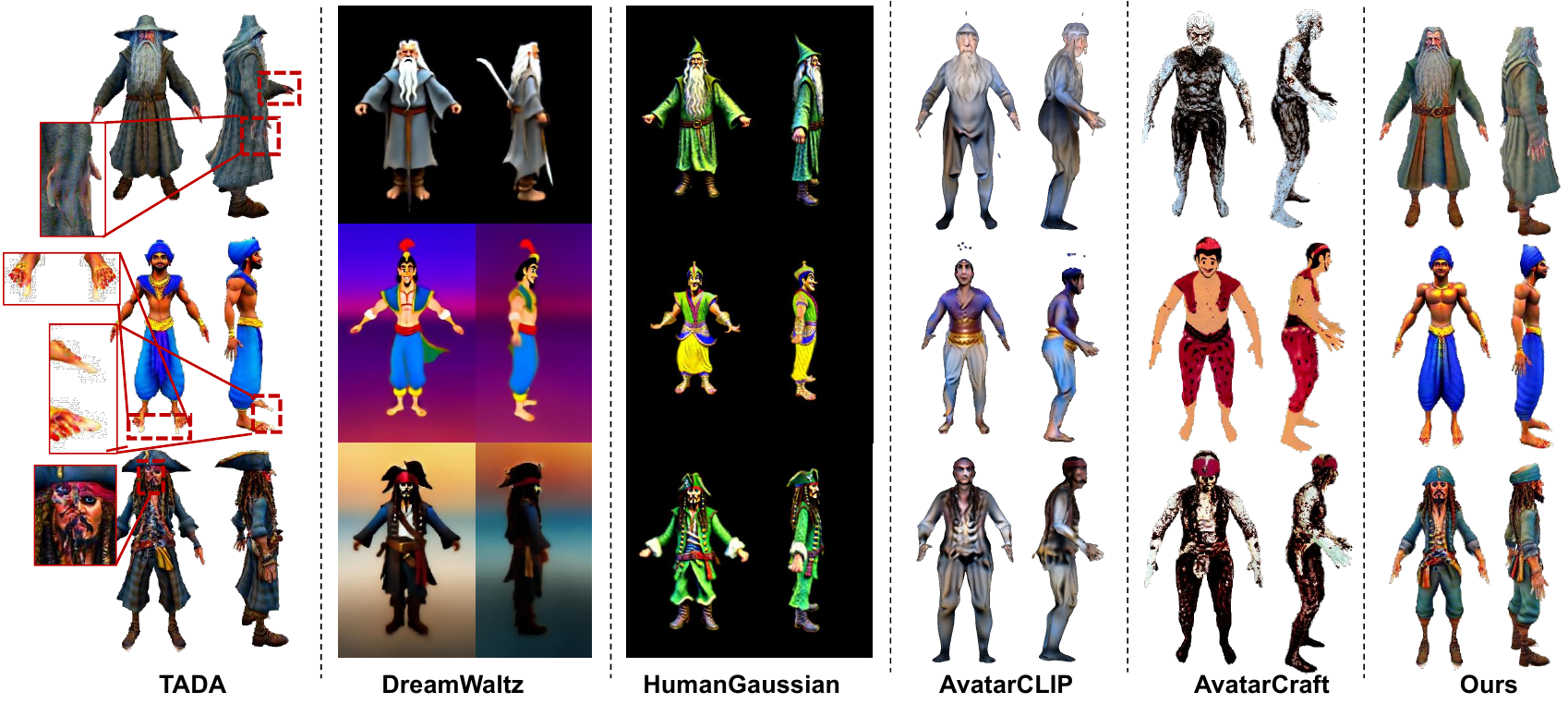}
\vspace{-1.0em}
\caption{Qualitative comparisons with SOTA text-to-avatar methods. The prompts (top → down) are ``Gandalf from The Lord of the Rings'', ``Aladdin in Aladdin'', and ``Captain Jack Sparrow from Pirates of the Caribbean''.}
\label{fig:sota1}
\vspace{-0.7em}
\end{figure*}

\begin{figure*}[]
\centering
\includegraphics[width=1.9\columnwidth]{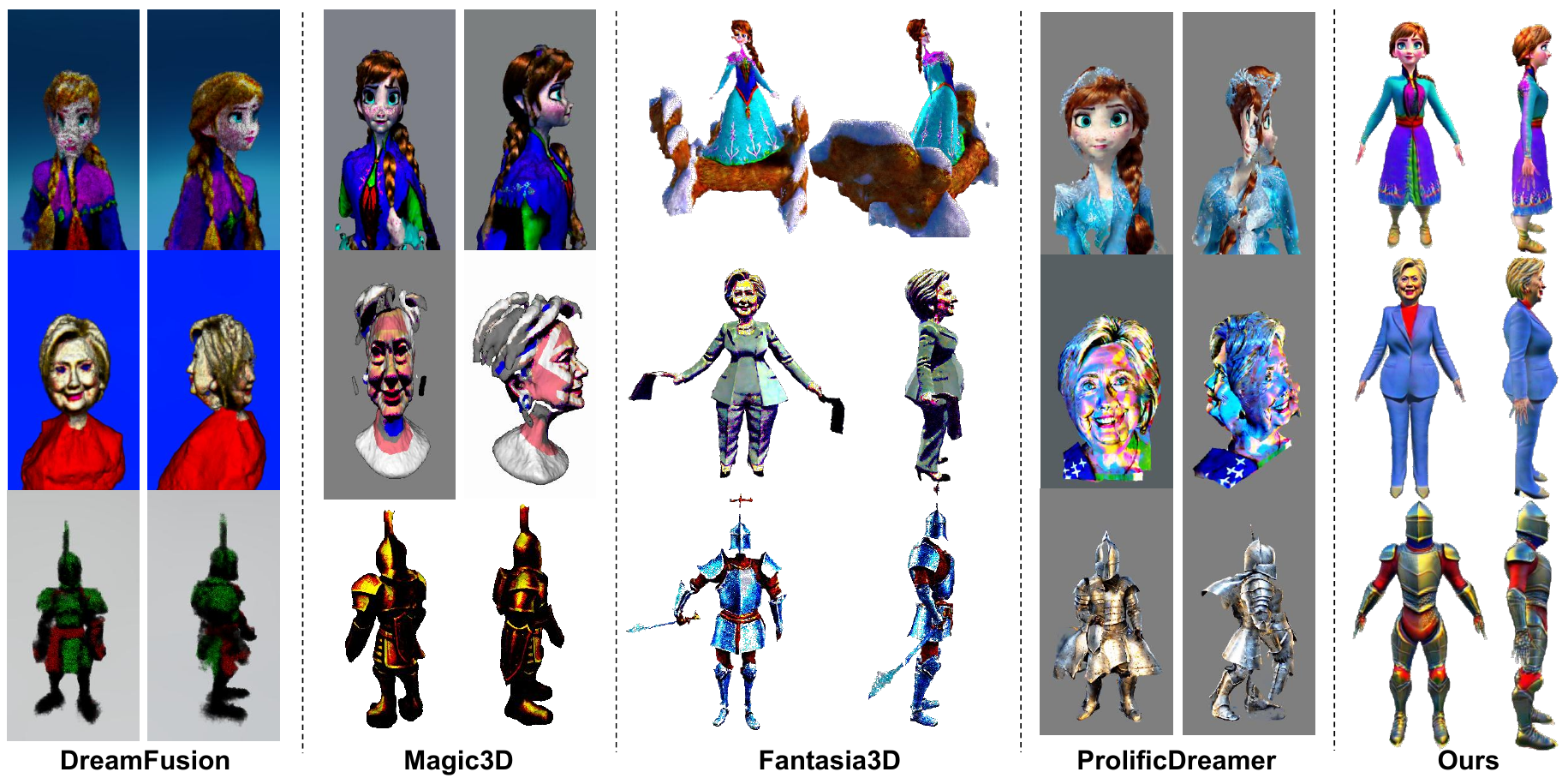}
\vspace{-1.0em}
\caption{Qualitative comparisons with SOTA text-to-3D methods. The prompts (top → down) are ``Anna in Frozen'', ``Hilary Clinton'', and ``Knight''.}
\label{fig:sota2}
\vspace{-0.8em}
\end{figure*}

\section{Experiments}
\label{sec:exper}

\subsection{Implementation Details}
Our experiments are conducted using a single Nvidia RTX 3090 GPU with 24GB of memory and the PyTorch library~\cite{paszke2019pytorch}.
The diffusion model employed in our implementation is the Stable Diffusion provided by HuggingFace Diffusers~\cite{von-platen-etal-2022-diffusers}.
During the training phase, we set the resolution of the rendered images to $800\times800$ pixels.
The resolution of the albedo map is $2048\times2048$ pixels.
The geometry modeling, appearance modeling, and animation refinement stages consist of 5000, 10000, and 5000 iterations, respectively.
We set the learning rates for the vertex offset $\psi_v$ and albedo map $\psi_a$ to 1e-4 and 5e-3, respectively.
Furthermore, we set the values of $\lambda_n$, $\lambda_v$, and $\lambda_a$ to 0.8, 0.1, and 0.1, respectively.
To enhance facial details, we employ a strategy where there is a 0.2 probability of rendering facial images for optimization during the training process, and a 0.8 probability of rendering full-body images for optimization.

\subsection{Comparison}

\begin{figure*}[]
\centering
\includegraphics[width=1.8\columnwidth]{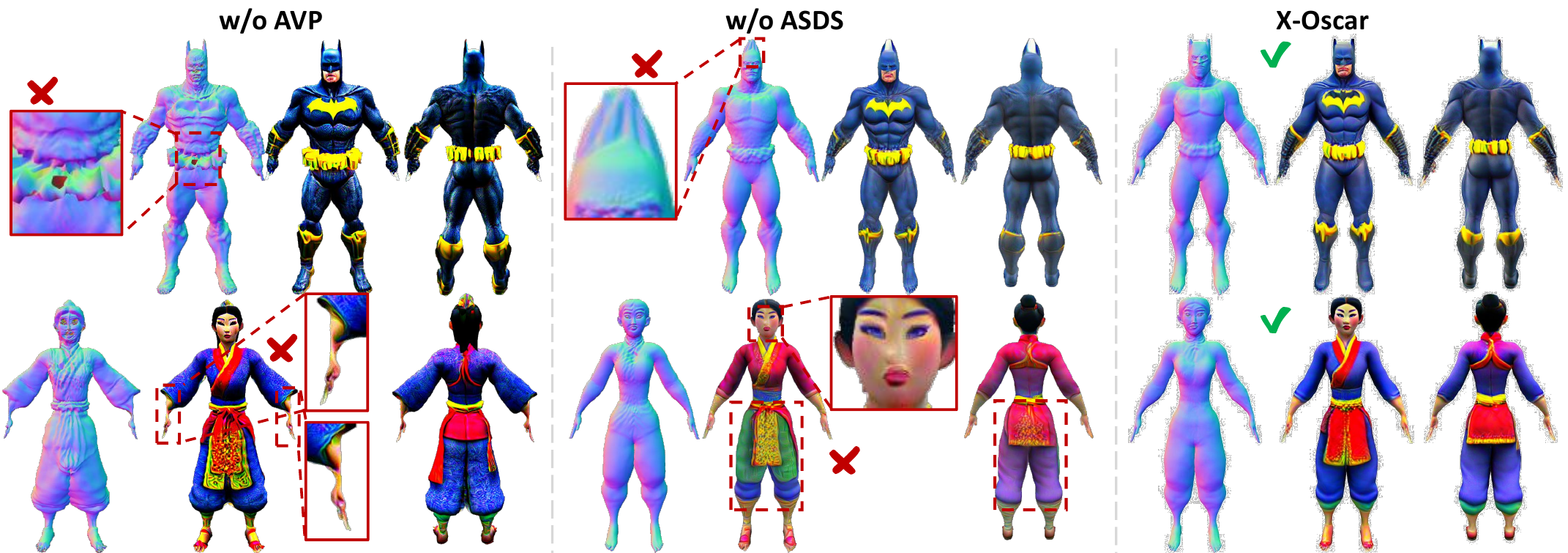}
\caption{Ablation study on the \moduleOneBig and \moduleTwoBig. The prompts (top → down) are ``Batman'', and ``Mulan''.}
\label{fig:ablation2}
\vspace{-1.5em}
\end{figure*}

\begin{figure}[]
\centering
\includegraphics[width=1.0\columnwidth]{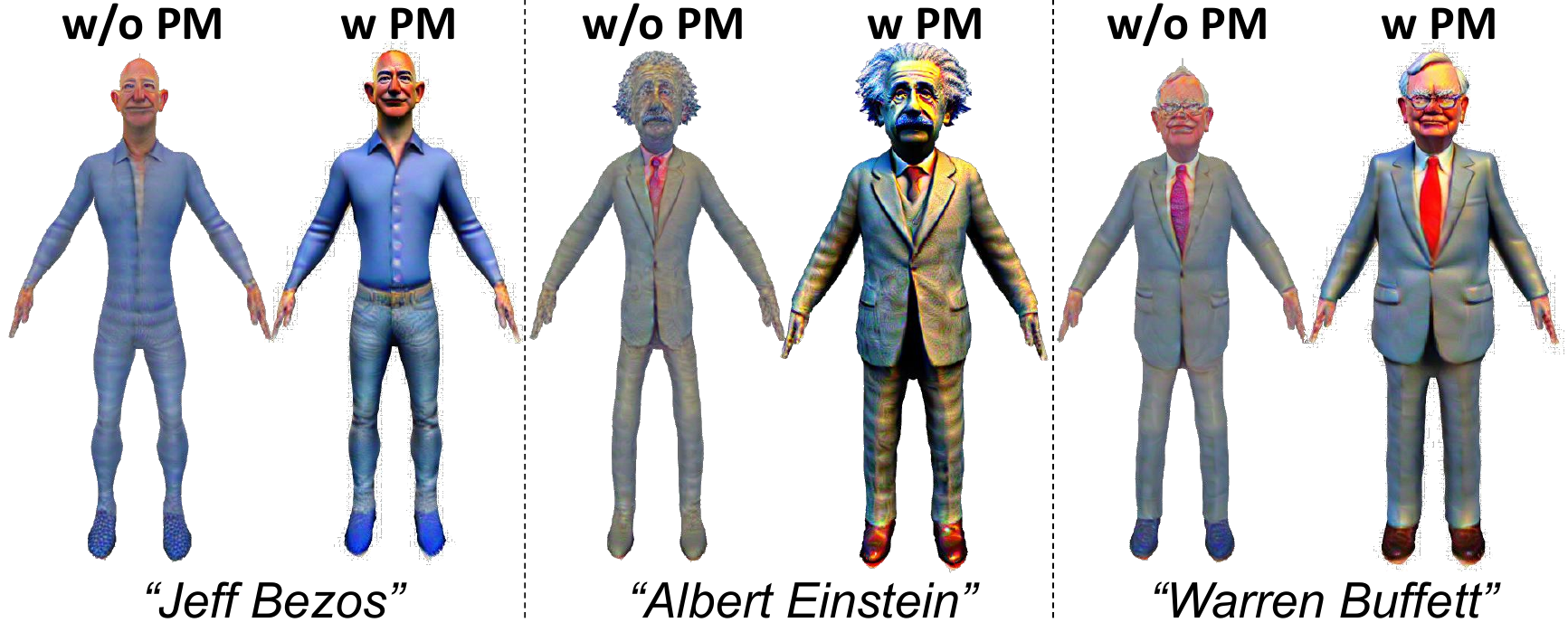}
\vspace{-1.0em}
\caption{Ablation study on progressive modeling. ``PM'' is short for ``progressive modeling''. ``w/o PM'' means that geometry, appearance, and animation are optimized together.}
\label{fig:ablation1}
\vspace{-1.5em}
\end{figure}

\textbf{Qualitative Comparison with Text-to-Avatar Methods.}
We present a comparative analysis of our methodology against five state-of-the-art (SOTA) baselines: TADA~\cite{liao2023tada}, DreamWaltz~\cite{huang2023dreamwaltz}, HumanGaussian~\cite{liu2023humangaussian}, AvatarCLIP~\cite{hong2022avatarclip}, and AvatarCraft~\cite{jiang2023avatarcraft}, as illustrated in \cref{fig:sota1}. 
We observe certain limitations in the geometry and texture of avatars generated by TADA, which we emphasize by enclosing them within a red box.
Furthermore, the outcomes produced by the other baselines exhibit issues such as blurriness and inconsistencies with the provided text.
In contrast, our proposed \modelname consistently generates high-quality avatars with intricate details.
Moreover, in addition to static avatars, \modelname is also capable of generating animatable avatars, as demonstrated in \cref{fig:intro}.

\textbf{Qualitative Comparison with Text-to-3D Methods.}
We also conduct a comparative analysis of \modelname with SOTA text-to-3D methods, namely DreamFusion~\cite{poole2022dreamfusion}, Magic3D~\cite{lin2023magic3d}, Fantasia3D~\cite{chen2023fantasia3d}, and ProlificDreamer~\cite{wang2023prolificdreamer}.
As shown in \cref{fig:sota2},  we observe evident limitations in the avatars generated by text-to-3D methods, including poor geometry and noisy texture.
Furthermore, owing to the absence of human prior knowledge, the avatars generated by text-to-3D methods lack flexibility and pose challenges in terms of animation.
In contrast, our proposed method excels in generating high-quality, animatable avatars.

\textbf{Quantitative Comparison.}
To assess \modelname quantitatively, we conduct user studies comparing its performance with SOTA text-to-3D content and text-to-avatar methods using the same prompts.
We randomly selected 40 prompts generated by ChatGPT for avatar creation, and the user studies involved 52 participants who provided subjective evaluations.
Participants rated the generated avatars based on three specific aspects: texture quality (Geo. Qua.), geometry quality (Tex. Qua.), and text consistency (Tex. Con.).
Scores range from 1 to 10, with higher scores indicating better quality.
As shown in \cref{tab:compare}, our method consistently outperforms all other methods across all evaluated aspects.
Additionally, we calculate similarity scores between the generated results and text prompts using CLIP~\cite{radford2021learning} and OpenCLIP~\cite{cherti2023reproducible} with different backbones.
Our method consistently achieves either the best or second-best results, demonstrating its ability to generate 3D avatars that are semantically consistent with the provided text prompts.

\subsection{Ablation Studies}
\textbf{Progressive Modeling.}
To evaluate the effectiveness of the progressive modeling paradigm in \modelname, we performed additional experiments by coupling the three training stages together.
The results shown in \cref{fig:ablation1} reveal a significant enhancement in the quality of geometry and appearance in the generated avatars when using the progressive modeling paradigm.
For example, consider the prompt ``Albert Einstein".
Without employing the progressive modeling approach, the generated avatar is limited to a rudimentary shape and color, lacking the intricate details necessary for recognizing Albert Einstein.
However, when employing the progressive modeling paradigm, we observe a remarkable improvement in the generated avatars.
%
%

\textbf{\moduleOneBig.}
To provide robust evidence of the impact of \moduleOneBigShort, we conducted comprehensive ablation studies by using specific parameters instead of distributions to represent avatars.
As depicted in \cref{fig:ablation2}, our observations strongly indicate that the omission of \moduleOneBigShort in \modelname can lead to an excessive optimization of geometry and appearance, as an effort to align the generated outputs with the text.
This subsequently leads to the problem of oversaturation.
Geometry oversaturation leads to topological overlay problems in the generated meshes, while appearance oversaturation results in avatars with exaggerated color contrast.
By integrating \moduleOneBigShort, we successfully tackle these issues, significantly improving the realism of both the geometry and appearance in the generated avatars.

\textbf{\moduleTwoBig.}
To investigate the impact of \moduleTwoBigShort, we conducted additional experiments by adding random Gaussian noise instead of avatar-aware noise to the rendered image for optimization.
As demonstrated in \cref{fig:ablation2}, the absence of \moduleTwoBigShort directly results in a noticeable decline in the overall quality of both the geometry and appearance of the generated avatars.
For instance, without \moduleTwoBigShort, two ears on Batman's head exhibit a geometric merging phenomenon. 
In the case of Mulan, the facial details become blurred and the colors on the front and back of the pants are inconsistent.
%

\section{Conclusion}
\label{conclusion}
This paper introduces \modelname, an advanced framework for generating high-quality, text-guided 3D animatable avatars. The framework incorporates three innovative designs to enhance avatar generation.
Firstly, we present a progressive modeling paradigm with clear and simple optimization objectives for each training stage.
Additionally, we propose \moduleOneBig (\moduleOneBigShort), which optimizes the distribution of avatars, addressing oversaturation.
Furthermore, we introduce \moduleTwoBig (\moduleTwoBigShort), leveraging avatar-aware denoising to enhance overall avatar quality.
Extensive experiments demonstrate the effectiveness of the proposed framework and modules.

\section*{Impact Statements}
This paper presents work whose goal is to advance the field of Machine Learning. There are many potential societal consequences of our work, none which we feel must be specifically highlighted here.


\bibliography{example_paper}
\bibliographystyle{icml2024}


\end{document}